\pdfoutput=1

\documentclass[11pt]{article}

\usepackage[]{ACL2023}
\usepackage{times}
\usepackage{latexsym}
\usepackage{spverbatim}
\usepackage{adjustbox}
\usepackage{fvextra}

\usepackage[T1]{fontenc}

\usepackage[utf8]{inputenc}

\usepackage{microtype}

\usepackage{multirow}
\usepackage[normalem]{ulem}
\useunder{\uline}{\ul}{}
\title{Graph-Augmented Relation Extraction Model with \\LLMs-Generated Support Document}
\author{
     Vicky Dong$^{*}$ \quad Hao Yu$^{*}$ \quad Yao Chen$^{*}$ \\
    School of Computer Science \\
    McGill University\\
    \texttt{\{vicky.dong,hao.yu2,yao.chen5\}@mail.mcgill.ca}\\
}

\begin{document}
\maketitle
\def\thefootnote{*}\footnotetext{These authors contributed equally to this work.}\def\thefootnote{\arabic{footnote}}

\begin{abstract}
This study introduces a novel approach to sentence-level relation extraction (RE) that integrates Graph Neural Networks (GNNs) with Large Language Models (LLMs) to generate contextually enriched support documents. By harnessing the power of LLMs to generate auxiliary information, our approach crafts an intricate graph representation of textual data. This graph is subsequently processed through a Graph Neural Network (GNN) to refine and enrich the embeddings associated with each entity ensuring a more nuanced and interconnected understanding of the data. This methodology addresses the limitations of traditional sentence-level RE models by incorporating broader contexts and leveraging inter-entity interactions, thereby improving the model's ability to capture complex relationships across sentences. Our experiments, conducted on the CrossRE dataset, demonstrate the effectiveness of our approach, with notable improvements in performance across various domains. The results underscore the potential of combining GNNs with LLM-generated context to advance the field of relation extraction.

\end{abstract}

\section{Introduction}
Among the fundamental tasks of NLP, relation extraction between entities from text holds paramount importance and plays an important role in diverse knowledge-dependent applications, such as machine reading comprehension \cite{Li2022} and molecular property prediction \cite{Zeng}. Identifying connections between entities within a single sentence poses several limitations:

Sentence-level (SL) models often miss broader contexts and struggle with relationships that extend beyond single sentences. They are less effective at identifying long-range dependencies, which is crucial for understanding complex sentence structures and distant entity relationships. 

In response to these challenges, drawing inspiration from the Graph Aggregation-and-Inference Network (GAIN) model proposed by \cite{zeng-etal-2020-gain}, we propose a graph-augmented relation model. The model aims to enrich word embeddings of entities by connecting their context and leveraging their interactions using an LLM-generated supplemental dataset via GCN.

\begin{figure*}
    \centering
    \includegraphics[width=1\linewidth]{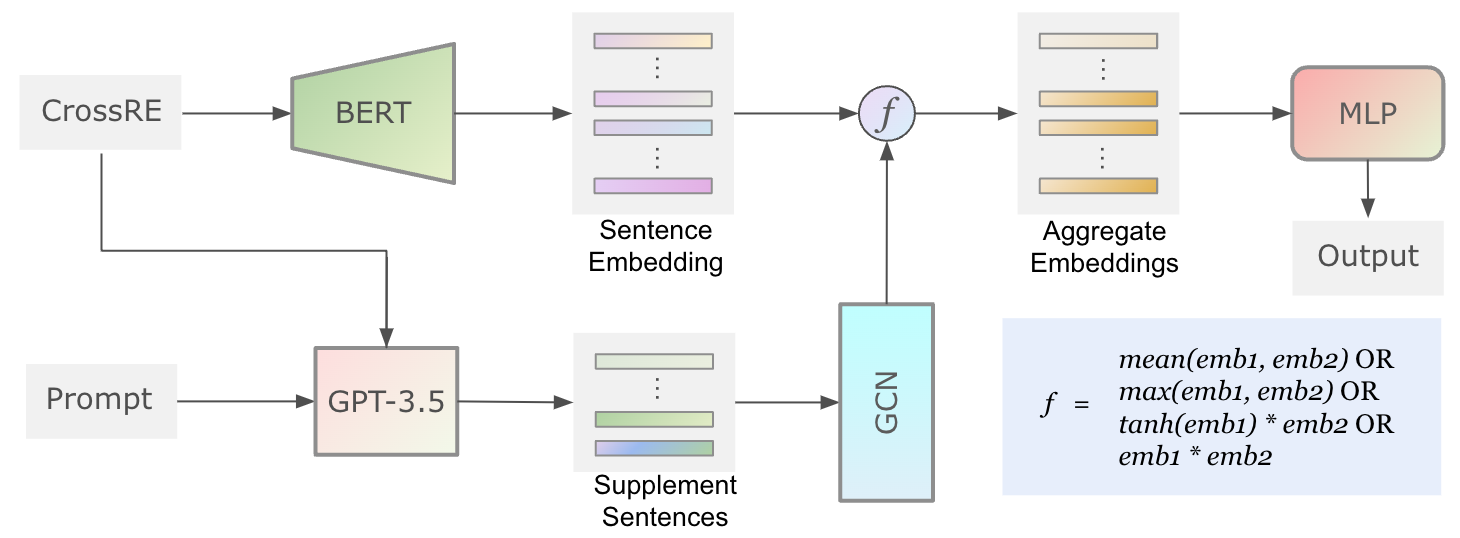}
    \caption{Model Overview}
    \label{fig:pipeline-demo}
\end{figure*}

To validate the performance and robustness of relation extraction models, particularly in capturing intricate relationships across multiple sentences, we will develop and evaluate our model using the cross-domain benchmark introduced by \cite{Bassignana2022}. Furthermore, an ablation study will be conducted to assess the impact of each component. We will measure the model's effectiveness using the Macro F1-Score and test its resilience against adversarial examples to ensure robust performance across diverse contexts and domains. The visualized pipeline is displayed in Figure \ref{fig:pipeline-demo}.

Our work includes the following contributions:
\begin{itemize}\vspace{-6pt}
    \item Using LLMs to generate supplement dataset for sentence-level relation extraction.\vspace{-6pt}
    \item Integrating embedding via the GNN module that takes an input of entity relation graph. \vspace{-6pt}
    \item Developing an evaluation framework  that systematically assesses the performance of GNN-aided models across various embeddings and base models.\vspace{-6pt} 
    \item Initiating domain-specific analysis for models, pinpointing unique challenges across diverse domains and outlining future directions to refine model training and evaluation.\vspace{-6pt} 
\end{itemize}

\section{Relative Work}
\label{sec:relative-work}
Current Sentence-level Relation Extraction (SLRE) methods, utilizing neural networks with architectures like CNNs, RNNs, and attention mechanisms\cite{zeng-etal-2014-relation,shen2016attention,alt2019improving}, have been significantly enhanced by the adoption of pre-trained models like BERT for their ability to leverage contextual information from extensive corpora. The introduction of a cross-domain benchmark for SLRE by \cite{Bassignana2022} marks an effort to address the gap in RE evaluation, which has traditionally been confined to in-domain scenarios. However, these SLRE methods often fail to capture relationships that span across multiple sentences due to loss of critical contextual information.
In contrast, Document-level Relation Extraction (DocRE) focuses on identifying relations between entities across entire documents \cite{Huang2021, Zhang2021, Xu2022, Ru2021}. 
Most existing DocRE approaches utilize dependency graphs to capture document-specific features, subsequently employing hierarchical inference networks \cite{Tang2020} or graph neural networks (GNN) 
 \cite{Zeng2020} for relational inference. However, these existing DocRE techniques could be improved by incorporating data augmentation strategies and enhancing the graph model structure.

To tackle sentence-level relation extraction problems, we have taken inspiration from document-level relation extraction and use graphs to establish relationships. In the following sections, we will delve deeper into these improvements.

\section{Dataset}
Our training, validating, and testing data are obtained from CrossRE \cite{Bassignana2022}, a sentence-level cross-domain relation extraction dataset introduced by Bassignana et al. This dataset is manually curated with hand-annotated relations spanning 17 relation types: {\small PART-OF, PHYSICAL, USAGE, ROLE, SOCIAL, GENERAL-AFFILIATION, COMPARE, TEMPORAL, ARTIFACT, ORIGIN, TOPIC, OPPOSITE, CAUSE-EFFECT, WIN-DEFEAT, TYPE-OF, NAMED, and RELATED-TO.} Some annotations are enriched with meta-data information, such as the explanation (Exp) for the choice of the assigned label, identification of syntax ambiguity (SA), and uncertainty of the annotator (UN).

\subsection{Supplemental Data}
To provide context for the sentences and entities in the sentences, ChatGPT 3.5 is leveraged to generate supplemental information. Without modifying the JSON format of the existing data, newly generated sentences are appended at the end of each data element. The prompt, \textit{"Generate some context for the given sentence: {Original sentence} while including the sentence in the paragraph generated. Keep the paragraph around 4 sentences."}, created a paragraph of 4 - 6 sentences.

\subsection{Examples}

Here is a sample data point from the CrossRE dataset:\vspace{-10pt}
\begin{small}
    
\begin{Verbatim}[breaklines]
Domain: Artificial Intelligence
Sentence: For many years starting from 1986, Miller directed the development of WordNet, a large computer-readable electronic reference usable in applications such as search engines.
Entities:
1. Miller (e1.researcher)
2. WordNet (e2.product)
3. search engines (e3.product)
\end{Verbatim}
\end{small}
\vspace{-12pt}

\begin{table}[h]
\centering
\begin{adjustbox}{width=0.4\textwidth}
\begin{tabular}{|c|c|c|c|c|c|}
\hline
Ent A & Ent B & Relation & Exp & SA & UN \\ \hline
$e_1$            & $e_2$                      & {\small ROLE}                     & -                     & -                      & -                      \\ \hline
$e_3$            & $e_2$                      & {\small USAGE}                     & -                     & -                      & {\small X}                      \\ \hline
\end{tabular}
\end{adjustbox}
\vspace{-15pt}
\end{table}

\paragraph{Entity}
Below is the table of entity types for each domain in the CrossRE dataset.\\

\begin{adjustbox}{width=0.48\textwidth}
\begin{tabular}{|l|p{6cm}|}
\hline
\textbf{Domain} & \textbf{Entity Types} \\ \hline
News & person, location, organization, misc \\ \hline
Politics & person, location, organization, misc \\ \hline
Natural Science & person, location, organization, misc, chemical, enzyme, protein, DNA, RNA, cell\_type, cell\_line \\ \hline
Music & person, location, organization, misc, instrument, genre, album, track \\ \hline
Literature & person, location, organization, misc, character, award, book \\ \hline
Artificial Intelligence & person, location, organization, misc, algorithm, model, task, method, conference, paper \\ \hline
\end{tabular}
\end{adjustbox}

\paragraph{Relationship}

This sample illustrates how the dataset captures the relationships between different entities within a sentence.
In addition, CrossRE allows for multi-label annotations \cite{Jiang2016}, meaning that each entity pair can be assigned to multiple relation types, except for the {\small RELATED-TO} label which is exclusive and has to be used when none of the other 16 labels fit the data. This enables more precise annotations that better represent the meaning expressed in the text.

\subsection{Cross Relation Entity (CrossRE)}
This dataset is derived from the CrossNER dataset \cite{Liu2020}, and contains six diverse text domains: news, politics, natural science, music, literature, and artificial intelligence, each with distinctive vocabularies, entity types, and relation distributions. The news domain is sourced from CoNLL-2003 \cite{Sang2003}, while the other five domains are collected from Wikipedia. The statistics of CrossRE are summarized in Table \ref{tab:crossre_stats}. The domains vary considerably in the amount and type of relations present in the data, posing a challenge for cross-domain generalization.

\begin{table}[h]
\centering
\begin{adjustbox}{width=0.48\textwidth}
\begin{tabular}{|l|c|c|c|c|}
\hline
\textbf{Domain} & \textbf{Train} & \textbf{Dev} & \textbf{Test} & \textbf{Relations} \\ \hline
News            & 164                      & 350                     & 400                     & 871                      \\ \hline
Literature      & 101                      & 350                     & 400                     & 3949                     \\ \hline
Natural Sciences& 103                      & 351                     & 400                     & 3088                     \\ \hline
Music           & 100                      & 350                     & 399                     & 4690                     \\ \hline
Politics        & 100                      & 400                     & 416                     & 3527                     \\ \hline
Artificial Intelligence              & 100                      & 350                     & 431                     & 2483                     \\ \hline
\end{tabular}
\end{adjustbox}
\caption{CrossRE Statistics: Number of Sentences and Relations Annotated Across Diverse Domains.}
\label{tab:crossre_stats}
\end{table}

\section{Methodology}
Our project introduces a novel approach to enhance sentence-level RE by incorporating document-level insights. Our method seamlessly integrates the Graph Neural Network (GNN) model, proficient at capturing complex inter-sentence relations and synthesizing information across an entire document, with the sentence-level model proposed in CrossRE. The objective is to establish a robust RE system capable of effectively handling both in-domain and out-of-distribution evaluation setups, while simultaneously improving the accuracy and context-awareness of relation extraction through the synergistic integration of these two approaches.

\subsection{LLMs for Support Document}
Research has illuminated the capacity of LLMs to be finely tuned through prompts, enabling them to generate text that is both contextually rich and highly relevant, drawing from initial sentences \cite{raffel2020exploring, brown2020language}. Inspired by the work of \cite{meng2023rapl}, our approach leverages the remarkable adaptability and capabilities of LLMs to craft tailored support documents for sentences within the CrossRE dataset. 

\subsection{Graph Neural Network Augmented}

GNNs are neural networks that operate on graph data. They have been developed for over a decade and recent developments have increased their capabilities and expressive power. \cite{SanchezLengeling2021} GNNs are used to process and understand graph-structured data, or information represented within a graph data structure. They provide a convenient way for node-level, edge-level, and graph-level prediction tasks.

To construct a comprehensive graph representation of these mini-documents, we iteratively identified words within each entity and any repeats in the supplemental document. 
\begin{itemize}
    \item A document \( D \) made up of \( N \) sentences:\\
    \centerline{\( D = \{s_i\}^N_{i=1} \)}\vspace{-8pt}
    \item Each sentence \( s_i \) consists of \( M \) words:\\
    \centerline{\( s_i = \{w_j\}^M_{j=1} \)}\vspace{-8pt}
    \item A set of entities \( E \) in the document:\\
    \centerline{\( E = \{e_i\}^P_{i=1} \),} each entity can be composed of more than one word.\vspace{-10pt}
    \item Each entity \( e_i \) has \( Q \) mentions:\\
    \centerline{\( e_i = \{m_j\}^Q_{j=1} \),} where \( m_j \) is any word within part of the entity in the document.\vspace{-10pt}
    \item Nodes within the same sentence or entity will be connected.\vspace{-3pt}
\end{itemize} 

The proposed architecture involves several steps. First, the sentences are tokenized and fed into a BERT model, which generates a contextualized embedding. Second, a graph representation is constructed using the word embeddings as node features and the adjacency matrix obtained from the previous steps. Finally, a Graph Convolutional Network (GCN) is used to aggregate the features of node neighbours.{
\setlength{\abovedisplayskip}{2pt} 
\setlength{\belowdisplayskip}{2pt}
\begin{equation}
    h_v^{(l+1)} = \sigma\left( \sum_{u \in \mathcal{N}(v)} \frac{1}{c_{vu}} W^{(l)} h_u^{(l)} \right)
\end{equation}}
where $h_v^{(l+1)}$ is the updated feature of node $v$ at layer $l+1$, $\mathcal{N}(v)$ is the set of neighbors of $v$, $c_{vu}$ is a normalization constant, $W^{(l)}$ is a weight matrix for layer $l$, and $\sigma$ is a non-linear activation function.

The graph structure represents the intricate interplay of entities and their relationships within each sentence. By extracting and embedding these sentence-level graphs, we create a rich tapestry of interconnected embeddings that encapsulate both the depth of individual sentences and the broader narrative flow of the document.

\subsection{Overall Architecture}
As introduced in the Figure \ref{fig:pipeline-demo}. The overall architecture would be that first the input document is vectorized and is consumed by an encoder to generate a contextualized embedding. Second, construct the graph representation using the word embeddings following the steps mentioned above. Third, apply Graph Attention Network (GAT) \cite{Brody2021} to aggregate mentions of each entity into a comprehensive graph embedding. Lastly, a classifier will perform link prediction between selected entities.

\section{Experiment Setup}
\subsection{Environment Setup}
All experiments were conducted on a single A100 80GB GPU on the \emph{conda} created environment. The details of the environment can be seen in the code repository later.

\subsection{Base Model}
\paragraph{BERT} (Bidirectional Encoder Representations from Transformers) \emph{bert-base-cased}: Used in the CrossRE \cite{Bassignana2022}.


\paragraph{RoBERTa (Robustly Optimized BERT Approach)} \emph{roberta-base}: RoBERTa is a variant of the BERT model. During training, the model was trained on a larger dataset and with a more effective training procedure that utilizes dynamic masking. This helps the model learn robust and generalizable representations of words.\cite{He2020}

\paragraph{DeBERTa-v3} \emph{deberta-v3-base}: DeBERTa-v3 improves the BERT and RoBERTa models using disentangled attention and enhanced mask decoder. In DeBERTa V3, the efficiency of DeBERTa was further improved using ELECTRA-Style \cite{Chi2021} pre-training with Gradient Disentangled Embedding Sharing. \cite{He2020}

\begin{table*}[ht!]
\centering
\begin{adjustbox}{width=0.85\textwidth}
\begin{tabular}{|c|c|cccccc|c|c|}
\hline
\multicolumn{1}{|l|}{\multirow{2}{*}{Base Model}} & \multirow{2}{*}{GNN Improvment} & \multicolumn{6}{c|}{Domain} & \multicolumn{1}{l|}{\multirow{2}{*}{Average}} & \multicolumn{1}{c|}{\multirow{2}{*}{\%$\triangle$}} \\
\multicolumn{1}{|l|}{} &  & \multicolumn{1}{l}{News} & \multicolumn{1}{l}{Politics} & \multicolumn{1}{l}{Science} & \multicolumn{1}{l}{Music} & \multicolumn{1}{l}{Literature} & \multicolumn{1}{l|}{AI} & \multicolumn{1}{l|}{} & \multicolumn{1}{l|}{} \\ \hline
\multirow{6}{*}{bert-base-cased} & None (Reported) & 10.10 & 20.00 & 21.60 & 30.70 & 36.40 & 20.80 & 23.30 & \multicolumn{1}{c|}{-} \\ \cline{2-10} 
 & None & 14.09 & {\ul 21.90} & {\ul 24.24} & \textbf{40.43} & 36.54 & {\ul 30.94} & 28.02 & 0.00 -\\
 & mean & {\ul 23.66} & {\ul 20.70} & {\ul 24.35} & {\ul 39.15} & \textbf{40.64} & 29.84 & 29.73 & +6.08 $\uparrow$\\
 & max & 23.73 & 18.83 & 23.93 & 37.54 & 36.20 & 29.38 & 28.27 & +0.88 $\uparrow$\\
 & tanh & \textbf{26.91} & 19.65 & \textbf{25.89} & {\ul 39.21} & 33.76 & \textbf{33.93} & \textbf{29.89} & +6.67 $\uparrow$ \\
 & times & 6.29 & 19.83 & 14.41 & 34.45 & 30.21 & 25.56 & 21.79 & -22.24 $\downarrow$\\ \hline
\multirow{5}{*}{roberta-base} & None & 5.25 & 18.92 & 21.01 & 20.94 & 35.13 & 23.94 & 20.87 & 0.00 -\\
 & mean & 5.46 & 16.68 & 22.66 & 19.08 & 29.68 & 24.01 & 19.59 & -6.10 $\downarrow$ \\
 & max & 5.46 & 17.20 & 23.76 & 32.50 & 26.34 & 27.50 & 22.13 &   +6.04 $\uparrow$\\
 & tanh & 5.46 & \textbf{25.73} & 5.67 & 19.31 & 27.41 & 27.06 & 18.44 & -11.63 $\downarrow$ \\
 & times & 5.46 & 17.69 & 8.69 & 17.34 & 25.30 & 20.13 & 15.77 & -24.43 $\downarrow$ \\ \hline
\multirow{5}{*}{deberta-v3-base} & None & 11.48 & 20.61 & 16.82 & 26.12 & {\ul 37.04} & 26.41 & 23.08 & 0.00 -\\
 & mean & 5.79 & 17.36 & 5.26 & 29.52 & 33.63 & {\ul 31.20} & 20.46 & -11.35 $\downarrow$ \\
 & max & 20.06 & 11.31 & 17.98 & 14.88 & 22.57 & 28.68 & 19.25 & -16.61 $\downarrow$ \\
 & tanh & 5.46 & 18.65 & 7.94 & 24.30 & {\ul 36.96} & 25.64 & 19.82 & -14.10 $\downarrow$ \\
 & times & {\ul 24.91} & 16.14 & 7.13 & 21.30 & 17.20 & 16.83 & 17.25 & -25.25 $\downarrow$ \\ \hline
\end{tabular}
\end{adjustbox}
\caption{Macro F1 Score Comparison of Different GNN-aided Models Across Various Embedding Methods. \emph{None (Reported)} represents the baseline according \cite{Bassignana2022}.}
\label{table:model-result}
\vspace{-5pt}
\end{table*}

\section{Experiment Result}
Our experimental results, as depicted in Table \ref{table:model-result}, offer a comprehensive overview of how different GNN-aided models, coupled with various embedding methods, perform across multiple domains. These results are pivotal in understanding the impact of integrating GNNs with different base models (\textit{bert-base-cased}, \textit{roberta-base}, and \textit{deberta-v3-base}) and employing distinct embedding techniques (mean, max, tanh, and times) on the Macro F1 score, a critical metric for evaluating the performance of relation extraction models.

The \textit{bert-base-cased} model, when augmented with the tanh embedding method, achieved a notable improvement, highlighting the efficacy of this particular combination. Conversely, the \textit{roberta-base} model saw a general improvement across all embedding methods, suggesting robust compatibility with the GNN module. The \textit{deberta-v3-base} model, however, presented a mixed outcome; while it showed an improvement with the times embedding method, it also indicated potential areas for optimization in other embedding techniques. Overall, the stronger the base model, the worse its performance in this task, as indicated by the average score of \textit{deberta-v3-base} and \textit{roberta-base}.

A key insight from these results is the domain-specific performance variability. For instance, the \emph{News} domain posed significant challenges, likely due to its imbalanced label distribution and the succinct nature of its content. This underscores the necessity for domain-specific strategies in model training and evaluation to address unique challenges and capitalize on domain-specific characteristics.

Moreover, the results underscore the critical role of the embedding method in optimizing model outcomes. The variance in performance across different base models and embedding techniques suggests that a more tailored approach to embedding selection could further enhance model performance. This observation opens up avenues for future research to explore the synergy between base models and embedding methods to optimize relation extraction tasks.

\section{Discussion}

The methodology explored in this study demonstrates considerable potential across various domains of the CrossRE dataset.The enhancement of performance across all foundational models, driven by the integration of the GNN module, underscores the effectiveness and value of this method. our findings confirm the pivotal importance of the embedding method in enhancing the efficacy of model outcomes, marking a significant area where we have achieved advancements. The observation that various embedding techniques achieve optimal results with different foundational models indicates that adopting a more  customized strategy for selecting embeddings could significantly boost model performance 

The challenge presented by the \emph{News} domain, characterized by its imbalanced label distribution and the concise nature of headlines and short reports, underscores the need for domain-specific considerations in both model training and evaluation. Furthermore, our study encountered challenges related to the use of GNNs, particularly the tendency for node representations to become overly similar (a phenomenon known as over-smoothing), and the static nature of GCNs, which are not well-suited to dynamic graphs that evolve over time.

To address these challenges and further advance the field, we propose several avenues for future research:
\begin{itemize}\vspace{-6pt} 
    \item Exploring a variety of GNN architectures that can mitigate the issue of over-smoothing and are better equipped to handle dynamically changing graphs.\vspace{-6pt} 
    \item Fine-tuning the application of adversarial training to better match the unique characteristics of different base models and the specific requirements of various data domains.\vspace{-6pt} 
    \item Developing strategies to address the imbalance in label distribution, possibly through the use of specialized models that are tailored to the content and structural peculiarities of specific domains.
\end{itemize}

\section{Conclusion}
In conclusion, this study presents an innovative approach to relation extraction that leverages the strengths of Graph Neural Networks (GNNs) and Large Language Models (LLMs) to enhance sentence-level analysis. Our experimental evaluations have demonstrated the promise of this model, showcasing its ability to improve performance across a variety of domains. As we continue to refine our methodology and explore new avenues for research, we are optimistic about the potential of our model to contribute significantly to the field of relation extraction.

\newpage


\bibliography{anthology,custom}

\begin{thebibliography}{23}
\expandafter\ifx\csname natexlab\endcsname\relax\def\natexlab#1{#1}\fi

\bibitem[{Alt et~al.(2019)Alt, H{\"u}bner, and Hennig}]{alt2019improving}
Christoph Alt, Marc H{\"u}bner, and Leonhard Hennig. 2019.
\newblock Improving relation extraction by pre-trained language representations.
\newblock \emph{arXiv preprint arXiv:1906.03088}.

\bibitem[{Bassignana and Plank(2022)}]{Bassignana2022}
Elisa Bassignana and Barbara Plank. 2022.
\newblock \href {https://aclanthology.org/2022.findings-emnlp.263} {{C}ross{RE}: A cross-domain dataset for relation extraction}.
\newblock In \emph{Findings of the Association for Computational Linguistics: EMNLP 2022}, pages 3592--3604, Abu Dhabi, United Arab Emirates. Association for Computational Linguistics.

\bibitem[{Brody et~al.(2021)Brody, Alon, and Yahav}]{Brody2021}
Shaked Brody, Uri Alon, and Eran Yahav. 2021.
\newblock \href {https://doi.org/10.48550/ARXIV.2105.14491} {How attentive are graph attention networks?}

\bibitem[{Brown et~al.(2020)Brown, Mann, Ryder, Subbiah, Kaplan, Dhariwal, Neelakantan, Shyam, Sastry, Askell et~al.}]{brown2020language}
Tom Brown, Benjamin Mann, Nick Ryder, Melanie Subbiah, Jared~D Kaplan, Prafulla Dhariwal, Arvind Neelakantan, Pranav Shyam, Girish Sastry, Amanda Askell, et~al. 2020.
\newblock Language models are few-shot learners.
\newblock \emph{Advances in neural information processing systems}, 33:1877--1901.

\bibitem[{Chi et~al.(2021)Chi, Huang, Dong, Ma, Zheng, Singhal, Bajaj, Song, Mao, Huang et~al.}]{Chi2021}
Zewen Chi, Shaohan Huang, Li~Dong, Shuming Ma, Bo~Zheng, Saksham Singhal, Payal Bajaj, Xia Song, Xian-Ling Mao, Heyan Huang, et~al. 2021.
\newblock Xlm-e: Cross-lingual language model pre-training via electra.
\newblock \emph{arXiv preprint arXiv:2106.16138}.

\bibitem[{He et~al.(2020)He, Liu, Gao, and Chen}]{He2020}
Pengcheng He, Xiaodong Liu, Jianfeng Gao, and Weizhu Chen. 2020.
\newblock \href {https://doi.org/10.48550/ARXIV.2006.03654} {Deberta: Decoding-enhanced bert with disentangled attention}.

\bibitem[{Huang et~al.(2021)Huang, Qi, Wang, Ma, and Huang}]{Huang2021}
Kevin Huang, Peng Qi, Guangtao Wang, Tengyu Ma, and Jing Huang. 2021.
\newblock \href {https://doi.org/10.18653/v1/2021.repl4nlp-1.30} {Entity and evidence guided document-level relation extraction}.
\newblock In \emph{Proceedings of the 6th Workshop on Representation Learning for NLP (RepL4NLP-2021)}, pages 307--315, Online. Association for Computational Linguistics.

\bibitem[{Jiang et~al.(2016)Jiang, Wang, Li, and Wang}]{Jiang2016}
Xiaotian Jiang, Quan Wang, Peng Li, and Bin Wang. 2016.
\newblock \href {https://aclanthology.org/C16-1139/} {Relation extraction with multi-instance multi-label convolutional neural networks}.
\newblock In \emph{{COLING} 2016, 26th International Conference on Computational Linguistics, Proceedings of the Conference: Technical Papers, December 11-16, 2016, Osaka, Japan}, pages 1471--1480. {ACL}.

\bibitem[{Li et~al.(2022)Li, Wong, Zhu, and Fung}]{Li2022}
Sirui Li, Kok~Kai Wong, Dengya Zhu, and Chun~Che Fung. 2022.
\newblock \href {https://doi.org/10.48550/ARXIV.2203.13570} {Improving question answering over knowledge graphs using graph summarization}.

\bibitem[{Liu et~al.(2020)Liu, Xu, Yu, Dai, Ji, Cahyawijaya, Madotto, and Fung}]{Liu2020}
Zihan Liu, Yan Xu, Tiezheng Yu, Wenliang Dai, Ziwei Ji, Samuel Cahyawijaya, Andrea Madotto, and Pascale Fung. 2020.
\newblock \href {http://arxiv.org/abs/2012.04373} {Crossner: Evaluating cross-domain named entity recognition}.

\bibitem[{Meng et~al.(2023)Meng, Hu, Liu, Li, Ma, Yang, and Wen}]{meng2023rapl}
Shiao Meng, Xuming Hu, Aiwei Liu, Shu'ang Li, Fukun Ma, Yawen Yang, and Lijie Wen. 2023.
\newblock Rapl: A relation-aware prototype learning approach for few-shot document-level relation extraction.
\newblock \emph{arXiv preprint arXiv:2310.15743}.

\bibitem[{Raffel et~al.(2020)Raffel, Shazeer, Roberts, Lee, Narang, Matena, Zhou, Li, and Liu}]{raffel2020exploring}
Colin Raffel, Noam Shazeer, Adam Roberts, Katherine Lee, Sharan Narang, Michael Matena, Yanqi Zhou, Wei Li, and Peter~J Liu. 2020.
\newblock Exploring the limits of transfer learning with a unified text-to-text transformer.
\newblock \emph{The Journal of Machine Learning Research}, 21(1):5485--5551.

\bibitem[{Ru et~al.(2021)Ru, Sun, Feng, Qiu, Zhou, Zhang, Yu, and Li}]{Ru2021}
Dongyu Ru, Changzhi Sun, Jiangtao Feng, Lin Qiu, Hao Zhou, Weinan Zhang, Yong Yu, and Lei Li. 2021.
\newblock \href {https://doi.org/10.18653/v1/2021.emnlp-main.95} {Learning logic rules for document-level relation extraction}.
\newblock In \emph{Proceedings of the 2021 Conference on Empirical Methods in Natural Language Processing}, pages 1239--1250, Online and Punta Cana, Dominican Republic. Association for Computational Linguistics.

\bibitem[{Sanchez-Lengeling et~al.(2021)Sanchez-Lengeling, Reif, Pearce, and Wiltschko}]{SanchezLengeling2021}
Benjamin Sanchez-Lengeling, Emily Reif, Adam Pearce, and Alexander~B. Wiltschko. 2021.
\newblock \href {https://doi.org/10.23915/distill.00033} {A gentle introduction to graph neural networks}.
\newblock \emph{Distill}.
\newblock Https://distill.pub/2021/gnn-intro.

\bibitem[{Sang and Meulder(2003)}]{Sang2003}
Erik F. Tjong~Kim Sang and Fien~De Meulder. 2003.
\newblock \href {https://doi.org/10.3115/1119176.1119195} {Introduction to the conll-2003 shared task}.

\bibitem[{Shen and Huang(2016)}]{shen2016attention}
Yatian Shen and Xuan-Jing Huang. 2016.
\newblock Attention-based convolutional neural network for semantic relation extraction.
\newblock In \emph{Proceedings of COLING 2016, the 26th International Conference on Computational Linguistics: Technical Papers}, pages 2526--2536.

\bibitem[{Tang et~al.(2020)Tang, Cao, Zhang, Cao, Fang, Wang, and Yin}]{Tang2020}
Hengzhu Tang, Yanan Cao, Zhenyu Zhang, Jiangxia Cao, Fang Fang, Shi Wang, and Pengfei Yin. 2020.
\newblock \href {https://doi.org/10.1007/978-3-030-47426-3_16} {{HIN}: Hierarchical inference network for document-level relation extraction}.
\newblock In \emph{Advances in Knowledge Discovery and Data Mining}, pages 197--209. Springer International Publishing.

\bibitem[{Xu et~al.(2022)Xu, Chen, Mou, and Zhao}]{Xu2022}
Wang Xu, Kehai Chen, Lili Mou, and Tiejun Zhao. 2022.
\newblock \href {https://doi.org/10.18653/v1/2022.naacl-main.212} {Document-level relation extraction with sentences importance estimation and focusing}.
\newblock In \emph{Proceedings of the 2022 Conference of the North American Chapter of the Association for Computational Linguistics: Human Language Technologies}, pages 2920--2929, Seattle, United States. Association for Computational Linguistics.

\bibitem[{Zeng et~al.(2014)Zeng, Liu, Lai, Zhou, and Zhao}]{zeng-etal-2014-relation}
Daojian Zeng, Kang Liu, Siwei Lai, Guangyou Zhou, and Jun Zhao. 2014.
\newblock \href {https://aclanthology.org/C14-1220} {Relation classification via convolutional deep neural network}.
\newblock In \emph{Proceedings of {COLING} 2014, the 25th International Conference on Computational Linguistics: Technical Papers}, pages 2335--2344, Dublin, Ireland. Dublin City University and Association for Computational Linguistics.

\bibitem[{Zeng et~al.(2020{\natexlab{a}})Zeng, Xu, Chang, and Li}]{zeng-etal-2020-gain}
Shuang Zeng, Runxin Xu, Baobao Chang, and Lei Li. 2020{\natexlab{a}}.
\newblock \href {https://www.aclweb.org/anthology/2020.emnlp-main.127} {Double graph based reasoning for document-level relation extraction}.
\newblock In \emph{Proceedings of the 2020 Conference on Empirical Methods in Natural Language Processing (EMNLP)}, pages 1630--1640. Association for Computational Linguistics.

\bibitem[{Zeng et~al.(2020{\natexlab{b}})Zeng, Xu, Chang, and Li}]{Zeng2020}
Shuang Zeng, Runxin Xu, Baobao Chang, and Lei Li. 2020{\natexlab{b}}.
\newblock \href {https://doi.org/10.48550/ARXIV.2009.13752} {Double graph based reasoning for document-level relation extraction}.

\bibitem[{Zeng et~al.()Zeng, Yao, Liu, and Sun}]{Zeng}
Zheni Zeng, Yuan Yao, Zhiyuan Liu, and Maosong Sun.
\newblock \href {https://doi.org/10.1038/s41467-022-28494-3} {A deep-learning system bridging molecule structure and biomedical text with comprehension comparable to human professionals}.
\newblock 13.

\bibitem[{Zhang et~al.(2021)Zhang, Chen, Xie, Deng, Tan, Chen, Huang, Si, and Chen}]{Zhang2021}
Ningyu Zhang, Xiang Chen, Xin Xie, Shumin Deng, Chuanqi Tan, Mosha Chen, Fei Huang, Luo Si, and Huajun Chen. 2021.
\newblock Document-level relation extraction as semantic segmentation.
\newblock \emph{arXiv preprint arXiv:2106.03618}.

\end{thebibliography}
\bibliographystyle{acl_natbib}
\end{document}